\documentclass[letterpaper, 10pt, conference]{ieeeconf}

\IEEEoverridecommandlockouts
\usepackage{amsmath}
\usepackage{amssymb}
\usepackage{graphicx}
\usepackage{algorithmic}
\usepackage{tcolorbox}
\usepackage{tabularx}
\usepackage{xcolor}
\usepackage{listings}
\usepackage{subcaption}
\usepackage{booktabs}
\usepackage{multirow}
\usepackage{cite}
\usepackage{url}

\tcbuselibrary{listings, breakable}

\newtcolorbox{userbox}{
  colback=gray!3!white,
  colframe=gray!40!black,
  title=Prompt Template,
  fonttitle=\bfseries,
  breakable
}

\lstdefinelanguage{PDDL}{
  morekeywords={define,domain,requirements,types,predicates,action,parameters,
                precondition,effect,init,goal,problem,when,not,and,or},
  sensitive=false
}

\lstdefinestyle{lispstyle}{
  language=PDDL,
  basicstyle=\ttfamily\small,
  keywordstyle=\bfseries\color{blue!60!black},
  commentstyle=\itshape\color{gray!80!black},
  columns=fullflexible,
  keepspaces=true,
  showstringspaces=false,
  frame=single,
  rulecolor=\color{black!20},
  backgroundcolor=\color{gray!3},
  frameround=tttt,
  breaklines=true,
  captionpos=b,
  xleftmargin=1em, xrightmargin=1em
}

\title{\LARGE \bf
SafeGen-LLM: Enhancing Safety Generalization in Task Planning for Robotic Systems
}

\author{
Jialiang Fan\textsuperscript{1},
Weizhe Xu\textsuperscript{1},
Mengyu Liu\textsuperscript{2},
Oleg Sokolsky\textsuperscript{3},
Insup Lee\textsuperscript{3}, and  Fanxin Kong\textsuperscript{1}\\
\emph{\textsuperscript{1}University of Notre Dame}
\emph{\textsuperscript{2}Washington State University }
\emph{\textsuperscript{3}University of Pennsylvania}\\
{\{jfan5, wxu3\}@nd.edu, mengyu.liu@wsu.edu, \{sokolsky, lee\}@seas.upenn.edu, fkong@nd.edu}
}

\begin{document}

\maketitle
\thispagestyle{empty}
\pagestyle{empty}

\begin{abstract}
    Safety-critical task planning in robotic systems remains challenging: classical planners suffer from poor scalability, Reinforcement Learning (RL)-based methods generalize poorly, and base Large Language Models (LLMs) cannot guarantee safety. 
    To address this gap, we propose safety-generalizable large language models, named SafeGen-LLM.
    SafeGen-LLM can not only enhance the safety satisfaction of task plans but also generalize well to novel safety properties in various domains. 
    We first construct a multi-domain Planning Domain Definition Language 3 (PDDL3) benchmark with explicit safety constraints.
    Then, we introduce a two-stage post-training framework: Supervised Fine-Tuning (SFT) on a constraint-compliant planning dataset to learn planning syntax and semantics, and Group Relative Policy Optimization (GRPO) guided by fine-grained reward machines derived from formal verification to enforce safety alignment and by curriculum learning to better handle complex tasks. 
    Extensive experiments show that SafeGen-LLM achieves strong safety generalization and outperforms frontier proprietary baselines across multi-domain planning tasks and multiple input formats (e.g., PDDLs and natural language).
\end{abstract}

\section{Introduction}

Robotic systems tightly integrate computation, communication, and physical processes, which have been widely deployed in safety-critical domains such as autonomous driving, industrial automation, and warehouse logistics. Unlike conventional computing systems, robotic systems interact directly with the physical world, where unsafe decisions may lead to irreversible consequences. For instance, in autonomous driving, a planning error can result in collisions; in industrial automation, unsafe operations may damage equipment or harm workers. These examples highlight that robotic task planning must go beyond efficiency and task completion: it must ensure safety under diverse and dynamic operating conditions.

Task planning is a core capability for robotic systems, endowing agents with the ability to organize and execute long-horizon tasks in constrained environments. Traditional task planners are predominantly search-based and operate on formal models expressed in the Planning Domain Definition Language (PDDL). Planners such as Fast Downward~\cite{helmert2006fast} and Metric-FF~\cite{hoffmann2001ff} employ heuristic search over symbolic state spaces to generate plans that can be verified against the underlying model.

% classical planner's limitation
However, classical planners exhibit fundamental limitations: (i) \emph{poor scalability}---solving time grows exponentially as problem complexity increases, and (ii) \emph{rigid input/output formats}---requiring substantial domain engineering and hand-crafted heuristics~\cite{geffner2013concise,ghallab2004automated}. When additional factors such as resource limits or safety constraints are introduced, these heuristics and search operators may no longer capture the relevant problem dynamics, leading to degraded performance or infeasible plans.

% RL planner's limitations
Learning-based planners attempt to alleviate these issues by using deep learning or reinforcement learning (RL) to learn heuristics or policies directly from data~\cite{wang2022ensuring,yu2021learning}.
Such methods can, in principle, incorporate safety constraints into the learning objective and generate safety-aware policies. However, RL-based planners also face critical limitations: (i) \emph{limited generalization}---trained policies typically handle only a single planning task \cite{packer2018assessing,kansky2017schema}, and (ii) \emph{high data and interaction cost}—achieving reliable performance often requires extensive environment interactions and large numbers of rollouts for training and evaluation~\cite{shivadekar2025artificial, dulac2020empirical}. As a result, their applicability in mission-critical robotic systems is fundamentally constrained.

Recently, Large Language Models (LLMs) have emerged as powerful general-purpose reasoning engines capable of capturing knowledge, following instructions, and generalizing across domains~\cite{cao2025large,plaat2024reasoning,liang2025ai}. LLMs offer high potential for robotic task planning because pretrained models can handle diverse and flexible input formats, from natural-language descriptions to symbolic PDDL specifications. Early studies show that LLMs can generate plausible plans from natural-language or symbolic inputs~\cite{yang2022automaton}, translate instructions into temporal logic specifications~\cite{pan2023data,van2024vernacopter}, or directly solve PDDL planning tasks under few-shot prompting~\cite{silver2024generalized}.

% current llm planner's limitation
However, base models without post-training exhibit critical deficiencies: they show low planning success rates and cannot guarantee safety. Without domain-specific safety knowledge and alignment with safety-critical decision preferences, LLMs may produce plans that are semantically incorrect, action-infeasible, or violate safety constraints, potentially leading to hazardous behaviors in real-world deployments.

These limitations raise a fundamental research question: how can we systematically align LLMs to generate safe task plans with strong \emph{safety generalization} across domains?
That is, aligned LLMs should not only enhance the safety satisfaction of task plans but also generalize to novel safety properties in various domains.

\begin{figure*}[htbp!]
  \centering
  \includegraphics[width=0.9\textwidth]{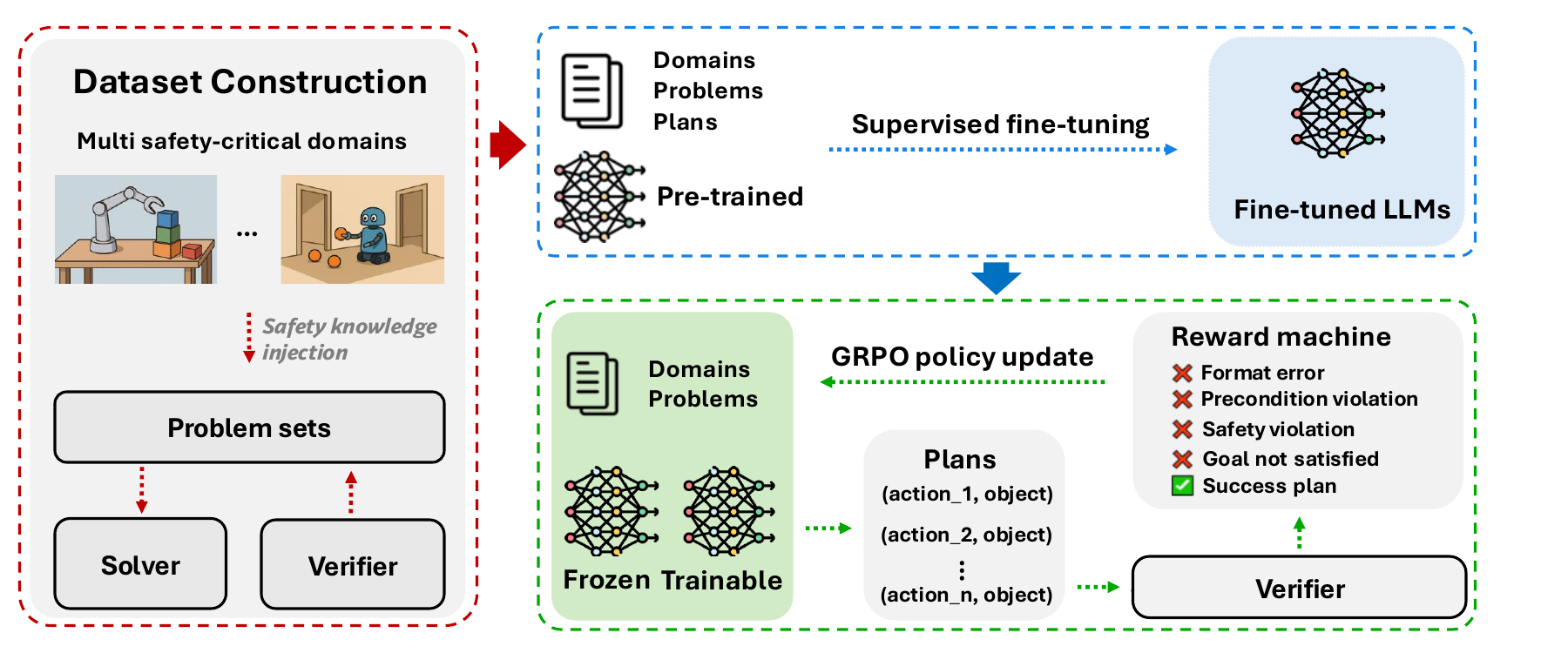}
  \caption{Overview of the proposed SafeGen-LLM framework. Stage I performs SFT on formally verified, safety-constrained plans. Stage II applies GRPO using fine-grained reward signals derived from formal verification to enforce safety alignment.}
  \label{fig:safe-gen-llm}
\end{figure*}

To address the question, we propose \emph{SafeGen-LLM} (Safety-Generalizable LLM), a post-training framework that enables large language models to perform safety-aware task planning in robotic systems by incorporating verifiable safety knowledge into the training process. As illustrated in Fig.~\ref{fig:safe-gen-llm}, the framework consists of three key components:
\begin{enumerate}
    \item \textbf{Safety-aware planning dataset construction}: We introduce a planning dataset that incorporates explicit safety constraints, enabling systematic training and evaluation of safety-aware planning models.
    \item \textbf{Stage I: Supervised Fine-Tuning (SFT)}: Built upon the proposed dataset, we apply SFT to enable the model to learn planning grammar and semantics.
    \item \textbf{Stage II: Group Relative Policy Optimization (GRPO) with fine-grained reward machines and curriculum learning}:  We devise a verifiable and fine-grained reward machine to guide GRPO ~\cite{shao2024deepseekmath} training, encouraging the LLM to achieve planning goals while maintaining safety alignment, and a curriculum learning approach to stabilize the training process and better handle complex planning tasks.
\end{enumerate}
% To ensure stable training, we adopt curriculum learning by progressively introducing planning problems with increasing complexity, improving both training stability and effectiveness.

Our main contributions are summarized as follows:
\begin{itemize}
    \item \textbf{A unified benchmark for safety-aware PDDL planning.}
    We introduce a dataset benchmark covering multiple robotics-inspired task-planning domains with explicitly defined  safety constraints, enabling systematic training and evaluation of safety compliance and generalization in planning tasks.

    \item \textbf{A systematic post-training framework for safe planning LLMs.}
    We propose a systematic two-stage post-training framework combining SFT and GRPO with fine-grained reward machines derived from formal verification. This approach improves the safety generalization of LLM-based planners, leading to higher planning success rates and more consistent safety compliance across multiple domains.

    \item \textbf{Cross-domain safety generalization with superior performance.}
    Through extensive experiments, we demonstrate that our trained models achieve strong planning performance across multiple domains and input formats with diverse safety constraints, effectively solving new problems for each domain while {outperforming frontier models with orders of magnitude more parameters} in safety-aware planning.
    Further, we show that safety can be guaranteed if SafeGen-LLM is integrated with LLM assurance frameworks such as SafePilot~\cite{xu2024assuring} while significantly increasing the efficiency of the planning process.
\end{itemize}

% The remainder of this paper is organized as follows.
% Section~\ref{sec:related} reviews related work on classical task planning, AI-based planners, and LLM-based planning.
% Section~\ref{sec:preliminaries} introduces preliminaries on classical planning and PDDL, and formulates the safety-aware and safety-generalizable planning problems.
% Section~\ref{IV} presents the proposed SafeGen-LLM framework.
% Section~\ref{V} describes the experimental setup and results.
% Finally, Section~\ref{VI} concludes with a discussion of limitations and future research directions.

\section{Related Work}\label{sec:related}
This section introduces the related work on classical task planning, AI-based planners, and LLM-based planning.

\subsection{Classical Task Planning}

Classical task planning is a mature area in automated reasoning, with planners such as
Fast Downward \cite{helmert2006fast}, Metric-FF \cite{hoffmann2001ff}, and OPTIC \cite{benton2012temporal},
solving symbolic planning problems by heuristic search over PDDL models.
Extensions including Hierarchical Task Networks (HTN) \cite{nau2005applications} and other model-based planners \cite{geffner2013concise} improve scalability through structural decomposition and domain-specific abstractions.

Despite their success, classical planners exhibit several well-recognized limitations. They rely heavily on expert-designed domain models and heuristic functions \cite{geffner2013concise}, making the modeling pipeline labor-intensive and brittle to environmental variations \cite{ghallab2004automated}. Moreover, as search-based methods, their computational cost often scales poorly with problem size, leading to rapidly increasing planning time as task complexity grows \cite{benton2012temporal}. These constraints motivate the shift toward more flexible,
data-driven planning paradigms.

\subsection{Learning-based Planners}
To address the rigidity of classical planners, a broad line of learning-driven
approaches has emerged. Early work focuses on learning heuristics or search guidance
from data \cite{yoon2008learning,shivashankar2015hierarchical}, enabling planners
to approximate expert reasoning without relying on manually crafted heuristics.
RL-based methods have also been explored to derive task policies from
environment interaction. However, RL-based planners often suffer from poor scalability due
to state and action space explosion, require extensive rollout-based data collection, and
exhibit limited generalization to novel goals or task configurations~ \cite{shivadekar2025artificial}. Neuro-symbolic \cite{acharya2023neurosymbolic} and hybrid approaches \cite{lyu2022towards} have explored integrating symbolic reasoning with learned components , but often remain domain-specific and require substantial manual engineering.

Overall, despite substantial progress, existing learning-based planners do not achieve
the joint requirements of (i) strong generalization across tasks and domains and
(ii) verifiable adherence to formal safety constraints—two properties that are essential
for robotic deployment.

\subsection{LLM-based Planning}
Recent work has investigated LLMs as planners that directly
generate or reason over plans for robotic and task planning.
For embodied agents, LLMs have demonstrated strong few-shot planning capabilities by
generating and updating plans grounded in environmental observations~\cite{song2023llm}.
LLMs have also been explored as generalized planners that synthesize domain-level planning
programs from a small number of training tasks, enabling efficient plan generation and
within-domain generalization~\cite{silver2024generalized}. To improve plan feasibility and correctness, several methods incorporate external
validation or refinement mechanisms.
ISR-LLM adopts an iterative self-refinement process with validation feedback to enhance
long-horizon planning performance~\cite{zhou2024isr}, while hybrid approaches combine LLM-based
reasoning with classical heuristic search planners for long-horizon or multi-agent
planning~\cite{zhang2025lamma}.
Recent benchmark studies further indicate that LLM planning performance can be
overestimated under simplistic settings, as introducing fine-grained constraints exposes
robustness and safety limitations~\cite{huang2025language}.

Overall, although recent LLM-based planners show promising planning performance, most approaches rely on frozen or lightly adapted models and external validation, and do not intrinsically learn safety-aware planning behavior or systematically generalize safety constraints across tasks and domains.

\section{Preliminaries}\label{sec:preliminaries}
This section briefly reviews classical planning and PDDL with safety constraints,
and then introduces the formal robotic task-planning model studied in this work.

\subsection{Classical Planning and PDDL}
In classical planning, a problem is defined as
$\Pi=\langle \mathcal{F}, \mathcal{A}, I, \mathcal{G} \rangle$,
where $\mathcal{F}$ denotes the set of fluents, $\mathcal{A}$ the action set,
$I\subseteq\mathcal{F}$ the initial state, and $\mathcal{G}$ the goal condition.
Each action $a\in\mathcal{A}$ is associated with preconditions $\mathrm{Pre}(a)$
and effects $\mathrm{Eff}(a)$.
A plan is a finite action sequence whose execution from $I$ leads to a state
satisfying $\mathcal{G}$.

Planning problems are commonly specified using
PDDL~\cite{aeronautiques1998pddl}, which provides a structured format
to describe actions, predicates, objects, initial states, and goals.
PDDL has become the de facto standard for representing classical planning benchmarks.

PDDL3 extends PDDL with constructs for specifying temporal constraints and preferences
over plans~\cite{gerevini2005plan}.
In particular, safety constraints can be encoded using the \texttt{:constraints}
field as temporal formulas that restrict all valid plans.
For example, the following constraint enforces that no block is ever placed on top
of a fragile block:
\begin{lstlisting}[style=lispstyle]
(:constraints
  (always
    (forall (?x - fragile-block ?y - block)
      (not (on ?y ?x)))))
\end{lstlisting}
In this work, we focus on hard safety constraints that must be satisfied by all
valid plans.

\subsection{Robotic Task Planning}
We formalize a robotic task-planning problem as a tuple
$\mathcal{P} = (\mathcal{S}, \mathcal{A}, s_0, \mathcal{G}, \mathcal{C})$,
where $\mathcal{S}$ denotes the (possibly hybrid) state space of the robotic system,
$\mathcal{A}$ is the set of available actions or control decisions,
$s_0 \in \mathcal{S}$ is the initial state,
$\mathcal{G}$ is the goal condition,
and $\mathcal{C}$ is a set of formal safety constraints that define the safe operating
region of the system.
Each constraint $C \in \mathcal{C}$ may specify either a state invariant that must hold
at every state along execution or a trajectory-level requirement capturing temporal
patterns such as ``eventually,'' ``until,'' or ``before.''

\paragraph{Safety-Aware Planning.}
A plan is defined as a finite sequence of actions
$\pi = [a_1, a_2, \dots, a_n]$
that induces a state trajectory
$s_0 \xrightarrow{a_1} s_1 \xrightarrow{a_2} \cdots \xrightarrow{a_n} s_n$.
The plan is \emph{goal-reaching} if the final state satisfies the goal condition,
i.e., $s_n \models \mathcal{G}$, and it is \emph{safe} if it respects all safety
constraints, denoted by $\pi \models \mathcal{C}$.
Safety-aware planning requires finding a plan that satisfies both conditions:
\begin{equation}
  \text{find } \pi \text{ s.t. } s_n \models \mathcal{G} \;\land\; \pi \models \mathcal{C}.
\end{equation}
While classical search-based planners enforce safety constraints explicitly during planning, LLM-based planners generate plans as sequences and do not inherently simulate state transitions against formal constraints. As a result, safety must be learned from data or feedback, or enforced through external
verification and correction mechanisms.

\paragraph{Safety-Generalizable Planning.}
Beyond solving individual planning problems, we require planners to generalize safety reasoning.
In safety-critical settings, this requirement extends to safety constraints themselves:
a planner should satisfy safety constraints in new problems or domains, a property
we refer to as \emph{safety generalizability}.
We distinguish two complementary forms of safety generalization.

\noindent\textbf{Cross-Problem Safety Generalizability.}
Fix a domain $D=(\mathcal{S}, \mathcal{A}, \mathcal{C})$ with shared state and action
semantics and a common set of safety constraints.
A model $\mathcal{M}$ exhibits cross-problem safety generalization if, for unseen problems
$\mathcal{P}_i \in D$ with different initial states and goals, it produces plans that are
both goal-reaching and safe:
\begin{equation}
  \forall \mathcal{P}_i \in D:\;
  \pi_i = \mathcal{M}(\mathcal{P}_i)
  \Rightarrow
  (s_n^i \models \mathcal{G}_i \;\land\; \pi_i \models \mathcal{C}).
\end{equation}
\noindent\textbf{Cross-Domain Safety Generalizability.}
Let $\{D_1, \dots, D_K\}$ denote a collection of planning domains, each with its own
state and action semantics and associated safety-constraint set.
A model $\mathcal{M}$ exhibits cross-domain safety generalizability if, for any domain
$D_j$ in this collection and any planning problem $\mathcal{P}_i \in D_j$, it produces
a plan that is both goal-reaching and safe:
\begin{equation}
  \forall D_j,\;
  \forall \mathcal{P}_i \in D_j:\;
  \pi_i = \mathcal{M}(\mathcal{P}_i)
  \Rightarrow
  (s_n^i \models \mathcal{G}_i \;\land\; \pi_i \models \mathcal{C}_j).
\end{equation}

Together, these requirements characterize the objective of safety-generalizable robotic
task planning studied in this work: the planner should produce plans that are safe by construction and sustain this property on unseen problems within and across domains.

\begin{table*}[t]
  \centering
  \caption{Brief introduction of the domains and their safety constraints.}
  \label{tab:domain-introduction}
  \begin{tabularx}{0.95\linewidth}{l|X|X}
    \hline
    \textbf{Domain} & \textbf{Description} & \textbf{Safety Constraints} \\
    \hline
    \textbf{Blocksworld} &
    The agent must rearrange stacked blocks into a desired configuration using pick-and-place actions. &
    Blocks must be stacked in a safe and stable order, e.g., a supporting block must be positioned before the blocks it supports. \\
    \hline
    \textbf{Ferry} &
    The agent controls a ferry that transports cars between locations. &
    The ferry must never be overloaded, and cars must safely reach their destinations before the ferry departs or returns. \\
    \hline
    \textbf{Grippers} &
    A robot with two grippers must pick up and deliver all objects to a target room. &
    Robots (or manipulators) must avoid operating simultaneously in the same narrow room, and certain grippers may be restricted for safety-critical objects. \\
    \hline
    \textbf{Spanner} &
    A worker must collect spanners and tighten all nuts across connected locations. &
    Bolts must be tightened in a safe sequence (e.g., foundation nuts before upper ones), and access to shared tools (e.g., entering the tool shed) is limited to prevent unsafe congestion. \\
    \hline
  \end{tabularx}
\end{table*}

\section{SafeGen-LLM}\label{IV}

Our framework consists of three stages: dataset construction (Section~\ref{subsec:dataset}), SFT (Section~\ref{subsec:sft}), and online RL via GRPO (Section~\ref{subsec:grpo}).

\subsection{Dataset Construction}\label{subsec:dataset}

We construct a unified safety-oriented dataset that is used for both SFT and as the problem pool for GRPO training.
The construction process involves three steps: domain selection and safety knowledge design, constrained problem generation and solving, and instruction--response formatting.

\noindent\textbf{Domain-specific safety knowledge design.}
We start by selecting four task-planning domains from the open-source PDDL2 problem generators~\cite{seipp2022pddl}.
The domains are chosen using the following criteria:
\begin{itemize}
  \item relevance to real-world robotic task planning;
  \item presence of safety-critical objects, locations, or actions;
  \item availability of problem instances with varying difficulty.
\end{itemize}
Based on these criteria, we select four representative domains: Blocksworld, Ferry, Grippers, and Spanner.
A brief overview of each domain and its associated safety constraints is given in Table~\ref{tab:domain-introduction}.
For each domain, we design additional domain-specific safety knowledge in the form of high-level constraints that mirror realistic robotic requirements (e.g., collision avoidance, load limits, safe ordering of operations).
These constraints are subsequently encoded in PDDL3 \texttt{:constraints} format and used to generate safety-compliant plan demonstrations.

\begin{figure}[h]
  \centering
  \includegraphics[width=1\linewidth]{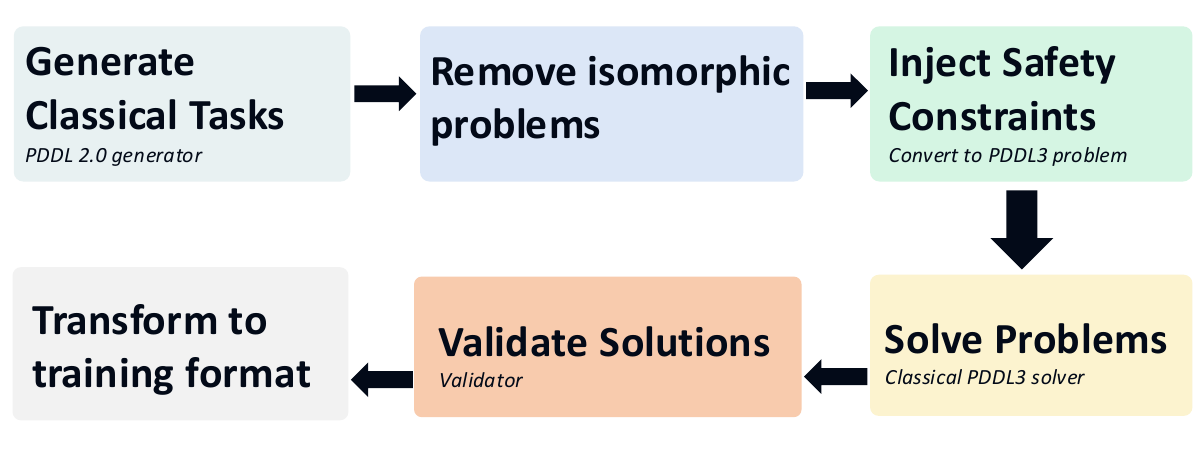}
  \caption{Pipeline for dataset construction.}
  \label{fig:dataset-construction}
\end{figure}

\noindent\textbf{Dataset construction pipeline.}
The pipeline for constructing the dataset is illustrated in Figure~\ref{fig:dataset-construction}.

First, we generate planning problems for each domain using the PDDL2 problem generators.
We remove isomorphic or trivially equivalent problems to reduce redundancy and employ a classical planner to ensure that the remaining problems are feasible.

Second, we encode the domain-specific safety constraints from Table~\ref{tab:domain-introduction} into PDDL3 \texttt{:constraints}.
We then use the temporal PDDL3 planner OPTIC~\cite{benton2012temporal} to solve the resulting constrained planning problems, and verify each candidate solution using the VAL tool~\cite{howey2004val}.
Only solutions that are successfully validated by VAL (i.e., respect both domain preconditions and encoded safety constraints) are retained.

Third, we convert each verified solution into an instruction--response pair.
The instruction consists of a natural-language prompt that includes the planning domain, problem instance, and (when applicable) safety constraints; the response is the corresponding validated plan.
Each problem is wrapped in a fixed instruction template that specifies the task format and expected output structure, as shown below.

\begin{userbox}
  You are a planning expert. Your task is to generate a \textbf{valid plan} for the given domain and problem.

  \texttt{DOMAIN:}
  \{\{domain\_content\}\}

  \texttt{PROBLEM:}
  \{\{problem\_content\}\}

  \textbf{Output Requirements:}
  \begin{itemize}
    \item Return \textbf{ONLY} the plan steps, one per line.
    \item Each line must follow the format: \texttt{(<ACTION\_NAME> <param1> <param2> ...)}.
    \item Use only objects defined in the \texttt{PROBLEM}.
    \item Do \textbf{NOT} include any explanations, comments, or headers.
    \item Do \textbf{NOT} output anything except the plan lines.
    \item The output must \textbf{NOT} contain natural language sentences.
    \item If the \texttt{PROBLEM} includes constraints, the plan must satisfy all of them; otherwise, solve as a standard goal-directed task.
    \item Ensure that all action preconditions hold and no constraints or invariants are violated at any step.
  \end{itemize}

  \textbf{Plan:}
\end{userbox}

This template explicitly instructs the model to behave as a planning expert and to produce strictly formatted action sequences.
By providing the domain and problem specifications in a structured form and enforcing strong output requirements (no extra text, strict syntax, constraint satisfaction), the template helps ensure that the generated plans remain syntactically correct, executable, and consistent with the safety constraints.

The resulting instruction--response pairs form the supervised dataset $\mathcal{D}_{\text{SFT}} = \{(x_i, y_i)\}_{i=1}^{N}$.
For SFT, these pairs are used directly as training examples.
For GRPO, the same problems serve as training prompts---the model generates its own candidate plans online---and the validated reference solutions $y_i$ are retained to compute the reference solution length $L_{\text{ref}}$ used in the progress-based reward function (Section~\ref{subsubsec:reward}).

\subsection{Supervised Fine-Tuning}\label{subsec:sft}

SFT adapts a pre-trained LLM to a specialized task by training on
task-specific instruction--response pairs.
Given a dataset $\mathcal{D}_{\text{SFT}}=\{(x_i, y_i)\}_{i=1}^{N}$, where $x_i$ is an
input prompt and $y_i$ is the target response, SFT optimizes the model parameters
$\theta$ by minimizing the negative log-likelihood:
\begin{equation}
\mathcal{L}_{\text{SFT}}(\theta)=
- \mathbb{E}_{(x,y)\sim\mathcal{D}_{\text{SFT}}}
\left[ \log \pi_{\theta}(y\mid x) \right],
\label{eq:sft_loss}
\end{equation}
where $\pi_{\theta}(y\mid x)$ denotes the probability of producing $y$ given input $x$.

In our framework, SFT serves to:
(i) encode domain and safety knowledge,
(ii) enforce syntactic and semantic plan validity, and
(iii) provide a strong initialization for subsequent RL via GRPO.

Given the constructed dataset $\mathcal{D}_{\text{SFT}}$ from Section~\ref{subsec:dataset}, we fine-tune the LLM using the standard SFT objective (Eq.~\ref{eq:sft_loss}).
By minimizing this objective, SFT encourages the model to reproduce planning behaviors that are syntactically valid, executable, and safety-compliant.
Beyond learning \emph{what} plan to output, SFT also enforces \emph{how} to output it, such as adhering to strict action syntax and avoiding natural-language explanations.
The resulting SFT model provides a capable initialization for the subsequent GRPO stage, which further refines safety compliance through online RL.

\subsection{Group Relative Policy Optimization}\label{subsec:grpo}

GRPO~\cite{shao2024deepseekmath} is an online RL algorithm designed to optimize language models using
verifiable, programmatic reward signals.
Given a prompt $x$, GRPO samples a group of $K$ candidate responses
$\{y_1, \dots, y_K\}$ from the current policy $\pi_\theta(\cdot \mid x)$.
Each response is evaluated by a reward function $r(x, y_i)$, and policy updates are
performed by comparing responses \emph{within the same group} rather than relying on
absolute reward values.

Specifically, GRPO defines a relative advantage for each response by normalizing
rewards within the group:
\begin{equation}
  A_i = r(x, y_i) - \frac{1}{K} \sum_{j=1}^{K} r(x, y_j),
\end{equation}
and optimizes the policy by maximizing:
\begin{equation}
  \mathcal{L}_{\mathrm{GRPO}}(\theta)
  = \mathbb{E}_{x,\, y_i \sim \pi_\theta}
  \big[ A_i \log \pi_\theta(y_i \mid x) \big].
\end{equation}

We adopt GRPO as the RL algorithm because it naturally integrates with our verifiable reward signal derived from the plan validator, while being significantly more lightweight than PPO by eliminating the need for a separate critic network.
Starting from the SFT-trained policy, GRPO encourages the model to generate safer and more successful plans through group-relative advantage estimation.
For each prompt, GRPO samples $K$ candidate plans from the current policy and evaluates each with a programmatic reward function.
We next describe the two key components of our GRPO training: the reward function that provides the learning signal, and the curriculum strategy that controls training difficulty.

\subsubsection{Reward design.}\label{subsubsec:reward}
We design a dense, hierarchical reward function based on automated plan verification using the VAL tool~\cite{howey2004val}.
For each generated plan, the validation pipeline classifies it into one of five ordered categories $\mathcal{C} = \{c_1, c_2, c_3, c_4, c_5\}$:
\begin{itemize}
  \item $c_1$: \textbf{Plan Format Error} --- syntactically invalid or unparseable;
  \item $c_2$: \textbf{Safety Constraint Violation} --- violates at least one PDDL3 safety constraint;
  \item $c_3$: \textbf{Precondition Violation} --- one or more action preconditions fail during execution;
  \item $c_4$: \textbf{Goal Not Satisfied} --- executes safely but fails to achieve the goal;
  \item $c_5$: \textbf{Success Plan} --- satisfies all safety constraints and achieves the goal.
\end{itemize}

\noindent\textbf{Hierarchical reward with progress-based interpolation.}
We assign each category $c_k$ a reward interval $[r_k^{-}, r_k^{+}]$ satisfying the \emph{strict separation} constraint:
\begin{equation}\label{eq:separation}
  r_1 \;\leq\; r_2^{-} \;<\; r_2^{+} \;\leq\; r_3^{-} \;<\; r_3^{+} \;\leq\; r_4^{-} \;<\; r_4^{+} \;\leq\; r_5,
\end{equation}
which guarantees that any plan in a more severe failure category always receives a lower reward than any plan in a less severe category, regardless of within-category progress.

For the two anchor categories, the reward is fixed: $r(x,y) = r_5$ for success and $r(x,y) = r_1$ for format errors.
For intermediate failure categories $c_k$ ($k \in \{2,3,4\}$), we interpolate using a \emph{progress function} $\rho_k(x, y) \in [0, 1]$:
\begin{equation}\label{eq:reward}
  r(x, y) = r_k^{-} + (r_k^{+} - r_k^{-}) \cdot \rho_k(x, y).
\end{equation}

The progress function is defined according to the failure type:
\begin{equation}\label{eq:progress}
\rho_k(x, y) =
\begin{cases}
  t_v / L_{\text{ref}} & k \in \{2, 3\}, \\[4pt]
  n_{\text{sat}} / n_{\text{total}} & k = 4,
\end{cases}
\end{equation}
where $t_v$ is the plan step at which the first violation occurs, $L_{\text{ref}}$ is the length of the ground-truth reference solution, and $n_{\text{sat}} / n_{\text{total}}$ is the fraction of satisfied goal predicates.

\noindent\textbf{Design rationale.}
Three principles guide the reward design.
\emph{First}, the strict separation (Eq.~\ref{eq:separation}) encodes a severity hierarchy (safety $>$ precondition $>$ goal), ensuring that the model learns to prioritize constraint compliance over goal achievement.
We rank safety violations as most severe because they correspond to explicitly specified domain-level invariants; precondition violations indicate local action misuse but do not necessarily imply unsafe states.
\emph{Second}, the progress-based interpolation (Eq.~\ref{eq:reward}) provides dense gradient signals within each failure category, encouraging the model to generate progressively longer valid prefixes rather than receiving a single uninformative penalty.
\emph{Third}, the denominator $L_{\text{ref}}$ in Eq.~\ref{eq:progress} uses the reference solution length rather than the length of the generated plan.
This reference-anchored normalization prevents \textbf{reward hacking}: without it, a model could inflate its progress ratio by generating shorter (and likely incorrect) plans.

\subsubsection{Curriculum learning.}\label{subsubsec:curriculum}
To improve training stability and sample efficiency, we adopt a curriculum learning strategy that progressively increases problem difficulty during GRPO training.

\noindent\textbf{Difficulty scoring.}
We define domain-specific difficulty scores based on structural parameters that correlate with planning complexity: $d = n^2$ for Blocksworld ($n$ blocks), $d = l \times c$ for Ferry ($l$ locations, $c$ cars), $d = n \times r \times o$ for Grippers ($n$ robots, $r$ rooms, $o$ objects), and $d = s \times n \times l$ for Spanner ($s$ spanners, $n$ nuts, $l$ locations).

\noindent\textbf{Three-level bucketing.}
Within each domain, problems are sorted by difficulty score and partitioned into three buckets using percentile thresholds: \emph{Easy} ($\leq$ 40th percentile), \emph{Medium} (40th--80th percentile), and \emph{Hard} ($>$ 80th percentile).
Bucketing is performed per domain since raw scores are not comparable across domains.

\noindent\textbf{Phased training schedule.}
Training is divided into three phases---early, mid, and late---with shifting sampling probabilities that progressively increase the proportion of harder problems.
The early phase builds foundational skills on easy problems; the mid phase transitions to a balanced distribution; the late phase emphasizes hard problems requiring complex multi-constraint reasoning.

\noindent\textbf{Domain-balanced batching.}
Each training batch contains an equal number of samples from every domain, preventing overfitting to any single domain and ensuring balanced exposure at every training step.

The combination of difficulty-based curriculum and domain-balanced batching enables GRPO to efficiently explore the solution space: easy problems build syntax and sequencing foundations, while progressively harder problems introduce complex constraint interactions that require deeper safety reasoning.

\section{Experiments}\label{V}
In this section, we empirically evaluate the proposed framework along four dimensions:
(i)~the scalability limitations of classical planners and the potential of LLM-based solvers,
(ii)~its cross-problem safety generalization within the same domain,
(iii)~its cross-domain safety generalization to multiple domains, and
(iv)~its effectiveness in a real-world robotic deployment on a physical robot arm.

\subsection{Experimental Setup}\label{subsec:experimental-setup}

We first describe the datasets, models, and training environment used in our experiments.

\noindent\textbf{Dataset construction.}
As introduced in Section~\ref{subsec:dataset}, we select four domains from the PDDL2 problem generators~\cite{seipp2022pddl}: \emph{Blocksworld}, \emph{Ferry}, \emph{Grippers}, and \emph{Spanner}.
For each domain, we generate $1{,}000$ planning problems with parameters ranging from simple to complex (e.g., different numbers of blocks, objects, or locations), of which $500$ are used for SFT and $500$ for GRPO training.
All problems are solved by the temporal planner OPTIC~\cite{benton2012temporal} and validated by VAL~\cite{howey2004val} to ensure correctness.
Importantly, these problems cannot be solved by the solutions of the corresponding PDDL2 benchmark problems without incorporating the additional safety constraints, which highlights the need for a safety-aware planner.
The test set consists of $50$ problems per domain.

\noindent\textbf{Model selection and training environment.}
We instantiate our framework with open-source LLMs available through the Unsloth library~\cite{unsloth} on Hugging Face~\cite{wolf2020transformers}.
Specifically, we consider three models: \textbf{Mistral-7B}\footnote{\texttt{unsloth/mistral-7b-instruct-v0.3-bnb-4bit}}, \textbf{Llama-8B}\footnote{\texttt{unsloth/Meta-Llama-3.1-8B-Instruct-bnb-4bit}}, and \textbf{Qwen3-14B}\footnote{\texttt{unsloth/Qwen3-14B-unsloth-bnb-4bit}}.
All models use quantized 4-bit variants throughout our experiments unless otherwise specified.
Training details, hyperparameters, reward function, and curriculum learning settings are provided in Appendix~\ref{app:training-details} and~\ref{app:reward-curriculum}.

\subsection{Running Time and Scalability Comparison}
We compare the runtime and success rates of GPT-5.2~\cite{openai2025gpt5} (LLM-based) with two classical planners---OPTIC~\cite{benton2012temporal} and Fast Downward~\cite{helmert2006fast}---on $48$ problems of increasing complexity in Blocksworld and Grippers, with a $300$s timeout.

\begin{figure}[!ht]
  \centering
  \begin{subfigure}[b]{\linewidth}
    \centering
    \includegraphics[width=\linewidth]{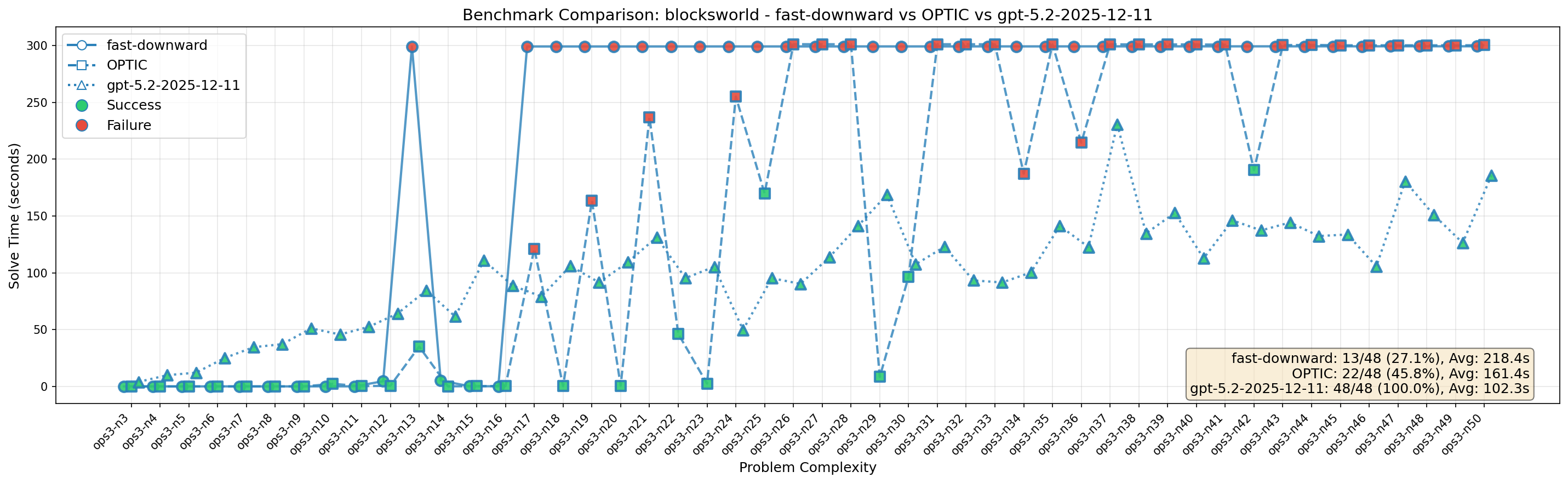}
    \caption{Blocksworld: GPT-5.2 100\% (48/48), avg 102.3s; OPTIC 45.8\% (22/48), avg 101.4s; Fast Downward 27.1\% (13/48), avg 218.4s.}
    \label{fig:benchmark-blocksworld}
  \end{subfigure}
  \vspace{0.3em}
  \begin{subfigure}[b]{\linewidth}
    \centering
    \includegraphics[width=\linewidth]{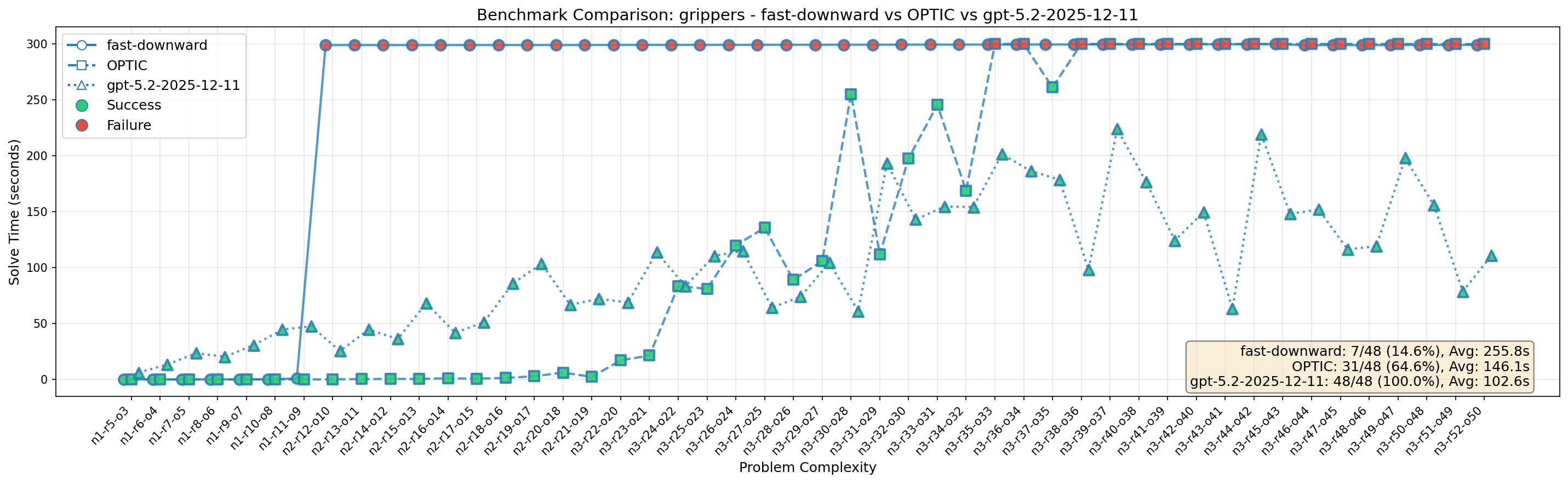}
    \caption{Grippers: GPT-5.2 100\% (48/48), avg 102.6s; OPTIC 64.6\% (31/48), avg 148.1s; Fast Downward 14.6\% (7/48), avg 255.8s.}
    \label{fig:benchmark-grippers}
  \end{subfigure}
  \caption{Three-way benchmark comparison: GPT-5.2 vs OPTIC vs Fast Downward. Green squares = success, red circles = failure.}
  \label{fig:benchmark-comparison}
\end{figure}

As shown in Figure~\ref{fig:benchmark-comparison}, GPT-5.2 achieves 100\% success in both domains with stable runtime ($\sim$102s), while both classical planners struggle as problem complexity grows. OPTIC solves 45.8\% of Blocksworld and 64.6\% of Grippers problems; Fast Downward performs worse at 27.1\% and 14.6\%, respectively, with the highest average runtimes. These results suggest that LLM-based solvers can alleviate scalability bottlenecks of classical planners.

We note that this experiment uses GPT-5.2 (via API) rather than our fine-tuned models, as the problem instances are \textbf{deliberately chosen to be highly complex} to stress-test classical planners, exceeding the capacity of our locally trained 7--14B parameter models under current GPU constraints.

\begin{figure}[h]
  \centering
  \includegraphics[width=0.9\linewidth]{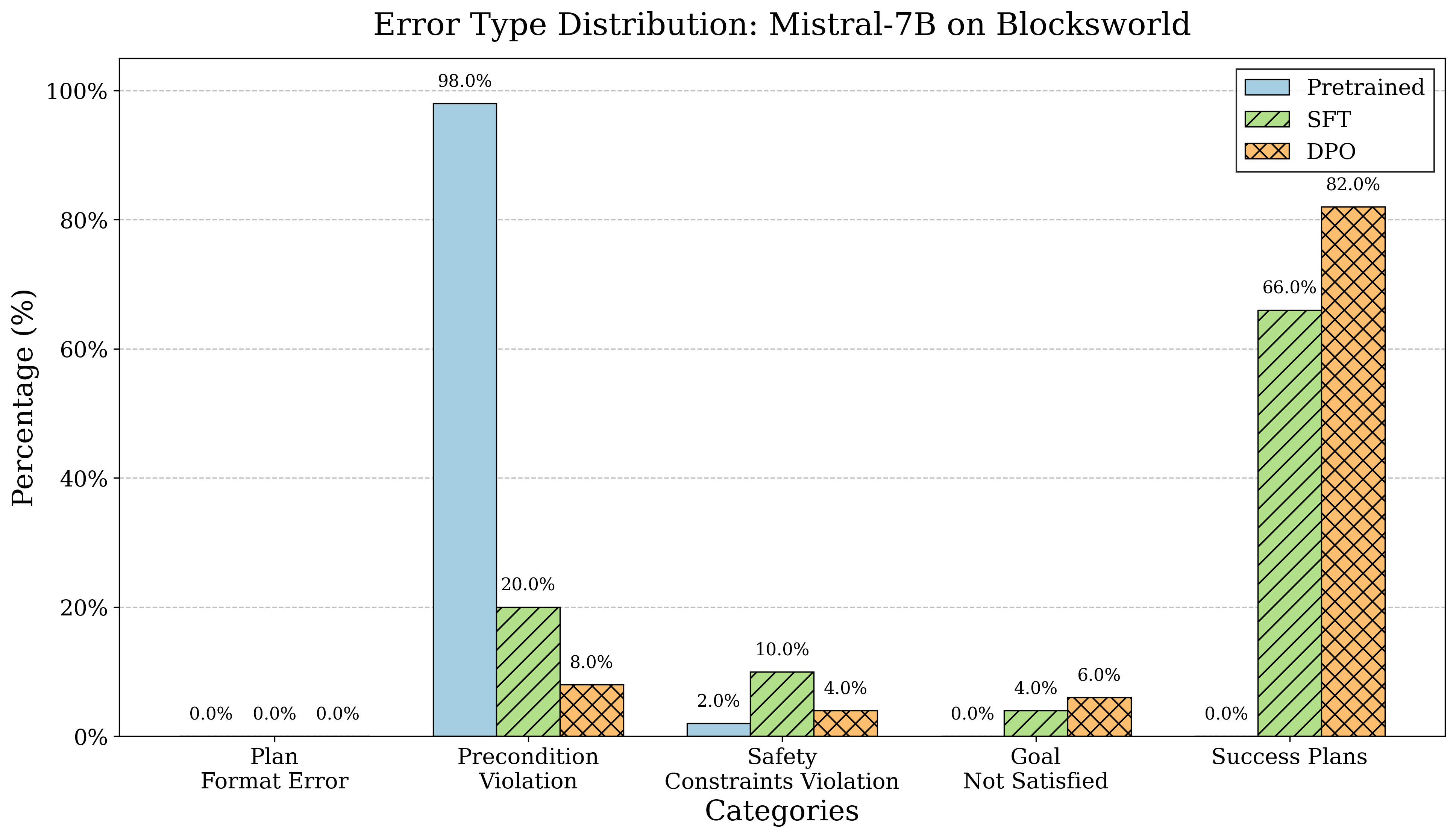}
  \caption{Error type distribution across training stages (Pretrained $\rightarrow$ SFT $\rightarrow$ GRPO) for Mistral-7B in the Blocksworld domain.}
  \label{fig:model-comparison}
\end{figure}

\begin{table*}[!t]
  \centering
  \caption{Error Type Percentages by Domain Across Training Stages}
  \label{tab:error_percentages}
  \resizebox{0.95\textwidth}{!}{%
  \begin{tabular}{l||ccc|ccc|ccc|ccc|ccc}
  \hline
  \textbf{Domain}
  & \multicolumn{3}{c|}{Success Plans}
  & \multicolumn{3}{c|}{Plan Format Error}
  & \multicolumn{3}{c|}{Precondition Violation}
  & \multicolumn{3}{c|}{Safety Constraint Violation}
  & \multicolumn{3}{c}{Goal Not Satisfied} \\
  \hline
  & \textbf{PT} & \textbf{SFT} & \textbf{GRPO}
  & \textbf{PT} & \textbf{SFT} & \textbf{GRPO}
  & \textbf{PT} & \textbf{SFT} & \textbf{GRPO}
  & \textbf{PT} & \textbf{SFT} & \textbf{GRPO}
  & \textbf{PT} & \textbf{SFT} & \textbf{GRPO} \\
  \hline
  \multicolumn{16}{l}{\textit{Qwen3-14B}} \\
  \hline
  Blocksworld & 2.0 & 70.0 & \textbf{88.0} & 90.0 & 0.0 & 0.0 & 6.0 & 24.0 & 12.0 & 0.0 & 2.0 & 0.0 & 2.0 & 4.0 & 0.0 \\
  Ferry       & 0.0 & 86.0 & \textbf{96.0} & 10.0 & 0.0 & 0.0 & 90.0 & 4.0 & 4.0 & 0.0 & 10.0 & 0.0 & 0.0 & 0.0 & 0.0 \\
  Grippers    & 2.0 & 92.0 & \textbf{98.0} & 0.0 & 0.0 & 0.0 & 46.0 & 4.0 & 0.0 & 46.0 & 4.0 & 2.0 & 6.0 & 0.0 & 0.0 \\
  Spanner     & 0.0 & 100.0 & \textbf{100.0} & 2.0 & 0.0 & 0.0 & 98.0 & 0.0 & 0.0 & 0.0 & 0.0 & 0.0 & 0.0 & 0.0 & 0.0 \\
  \hline
  \multicolumn{16}{l}{\textit{Llama-8B}} \\
  \hline
  Blocksworld & 0.0 & 50.0 & \textbf{78.0} & 78.0 & 0.0 & 0.0 & 22.0 & 38.0 & 16.0 & 0.0 & 2.0 & 2.0 & 0.0 & 10.0 & 4.0 \\
  Ferry       & 0.0 & 80.0 & \textbf{88.0} & 2.0 & 0.0 & 0.0 & 98.0 & 4.0 & 8.0 & 0.0 & 16.0 & 4.0 & 0.0 & 0.0 & 0.0 \\
  Grippers    & 0.0 & 66.0 & \textbf{82.0} & 42.0 & 0.0 & 0.0 & 42.0 & 22.0 & 14.0 & 16.0 & 8.0 & 2.0 & 0.0 & 4.0 & 2.0 \\
  Spanner     & 0.0 & 92.0 & \textbf{94.0} & 100.0 & 0.0 & 0.0 & 0.0 & 8.0 & 6.0 & 0.0 & 0.0 & 0.0 & 0.0 & 0.0 & 0.0 \\
  \hline
  \end{tabular}
  }
  \vspace{0.5em}

  \footnotesize{PT = Pretrained; SFT = Supervised Fine-Tuning; GRPO = Group Relative Policy Optimization. Values in \%.}
\end{table*}

\begin{table*}[!t]
  \centering
  \caption{Performance comparison: Symbolic (PDDL3) vs Natural Language vs JSON input. All values in \%.}
  \label{tab:input_format_comparison}
  \resizebox{0.95\textwidth}{!}{%
  \begin{tabular}{l||ccc|ccc|ccc|ccc|ccc}
  \hline
  \textbf{Domain}
  & \multicolumn{3}{c|}{Success $\uparrow$}
  & \multicolumn{3}{c|}{Format Error}
  & \multicolumn{3}{c|}{Precondition Violation}
  & \multicolumn{3}{c|}{Safety Constraint Violation}
  & \multicolumn{3}{c}{Goal Not Satisfied} \\
  \hline
  & \textbf{PDDL3} & \textbf{NL} & \textbf{JSON}
  & \textbf{PDDL3} & \textbf{NL} & \textbf{JSON}
  & \textbf{PDDL3} & \textbf{NL} & \textbf{JSON}
  & \textbf{PDDL3} & \textbf{NL} & \textbf{JSON}
  & \textbf{PDDL3} & \textbf{NL} & \textbf{JSON} \\
  \hline
  Blocksworld & \textbf{88.0} & 74.0 & 80.0 & \textbf{0.0} & \textbf{0.0} & \textbf{0.0} & 12.0 & 18.0 & \textbf{8.0} & \textbf{0.0} & 8.0 & 8.0 & \textbf{0.0} & \textbf{0.0} & 4.0 \\
  Ferry       & \textbf{96.0} & 90.0 & \textbf{96.0} & \textbf{0.0} & \textbf{0.0} & \textbf{0.0} & 4.0 & 6.0 & \textbf{2.0} & \textbf{0.0} & 4.0 & 2.0 & \textbf{0.0} & \textbf{0.0} & \textbf{0.0} \\
  Grippers    & \textbf{98.0} & 82.0 & \textbf{98.0} & \textbf{0.0} & \textbf{0.0} & \textbf{0.0} & \textbf{0.0} & 14.0 & 2.0 & 2.0 & 4.0 & \textbf{0.0} & \textbf{0.0} & \textbf{0.0} & \textbf{0.0} \\
  Spanner     & \textbf{100.0} & 90.0 & 96.0 & \textbf{0.0} & 2.0 & \textbf{0.0} & \textbf{0.0} & 8.0 & 4.0 & \textbf{0.0} & \textbf{0.0} & \textbf{0.0} & \textbf{0.0} & \textbf{0.0} & \textbf{0.0} \\
  \hline
  \textbf{Average} & \textbf{95.5} & 84.0 & 92.5 & \textbf{0.0} & 0.5 & \textbf{0.0} & \textbf{4.0} & 11.5 & \textbf{4.0} & \textbf{0.5} & 4.0 & 2.5 & \textbf{0.0} & \textbf{0.0} & 1.0 \\
  \hline
  \end{tabular}
  }
\end{table*}

\subsection{Cross-Problem Safety Generalizability}
We begin with cross-problem safety generalization, i.e., whether a model trained on planning problems within a given domain can solve previously unseen problems with safety constraints in the \emph{same} domain.

We use the Blocksworld domain as a representative example and fine-tune Mistral-7B through our two-stage pipeline (SFT followed by GRPO).
Figure~\ref{fig:model-comparison} compares the error type distribution across three stages: Pretrained, SFT, and GRPO.

We summarize the key observations as follows:
\begin{enumerate}
  \item \textbf{Dramatic reduction in precondition violations.}
  The precondition violation rate decreases from 98\% (Pretrained) to 20\% (SFT) to 8\% (GRPO).
  This indicates that both training stages progressively internalize domain-specific action semantics.

  \item \textbf{Progressive improvement in safety compliance.}
  Safety constraint violations decrease from 10\% (SFT) to 4\% (GRPO).
  Note that safety violations are computed only for parsable plans that can be executed by the validator up to the first failure; the low 2\% rate of the Pretrained model does not reflect genuine safety awareness, as most of its plans fail before reaching the safety-checking stage due to format or precondition errors.

  \item \textbf{Substantial success rate improvement.}
  The success rate improves dramatically: from 0\% (Pretrained) to 66\% (SFT) to 82\% (GRPO).
  This demonstrates that our two-stage training pipeline enables the model to generate plans that both achieve goals and satisfy safety constraints.
\end{enumerate}

These results demonstrate that our framework achieves substantial improvements in safety-aware planning through the combination of SFT and GRPO.

\subsection{Cross-domain Safety Generalizability}

We next evaluate whether the proposed framework generalizes across both model scales and planning domains.
To this end, we apply the same two-stage training pipeline (SFT followed by GRPO) to two additional models---Qwen3-14B and Llama-8B---and train each on all four domains simultaneously using domain-balanced batching and curriculum learning.
Both models follow the identical training protocol and hyperparameters described in Section~\ref{subsec:experimental-setup} and Appendix~\ref{app:training-details}.

Table~\ref{tab:error_percentages} summarizes the error-type distributions across all training stages---Pretrained (PT), SFT, and GRPO---for both models.
We highlight several key findings:

\begin{enumerate}
    \item \textbf{Pretrained models produce mostly format and precondition errors.}
    Without fine-tuning, both models fail almost entirely: success rates are 0--2\%.
    Format errors range from 2--100\% and precondition violations from 6--98\%, as pretrained models have never encountered PDDL plan syntax.

    \item \textbf{SFT eliminates format errors and dramatically improves success.}
    After SFT, format errors drop to 0\% for both models across all domains.
    Success rates jump to 70--100\% for Qwen3-14B and 50--92\% for Llama-8B, demonstrating that supervised learning effectively teaches plan structure and domain semantics.

    \item \textbf{GRPO further improves success while reducing safety violations.}
    GRPO raises Qwen3-14B success rates to 88--100\% and Llama-8B to 78--94\%.
    Safety violations are nearly eliminated, reaching 0--2\% for Qwen3-14B and 0--4\% for Llama-8B.
    This confirms that online RL with verifiable rewards effectively teaches models to prioritize safety compliance.

    \item \textbf{Consistent trends across model scales.}
    Both models exhibit the same progression: PT $\rightarrow$ SFT eliminates format errors and builds domain knowledge, while SFT $\rightarrow$ GRPO refines safety compliance and boosts success.
    Qwen3-14B achieves higher absolute performance, as expected given its larger capacity, but the smaller Llama-8B still attains strong results (e.g., 88\% on Ferry, 82\% on Grippers), demonstrating the framework's generalizability across model scales.
\end{enumerate}

These results show that the proposed training pipeline---SFT followed by GRPO---enables progressive improvements in both safety and success across different model scales.
The combination of curriculum learning and verifiable rewards allows models to explore and learn from their own generated plans, leading to robust safety-aware planning behavior.

\begin{figure}[t]
  \centering
  \includegraphics[width=\linewidth]{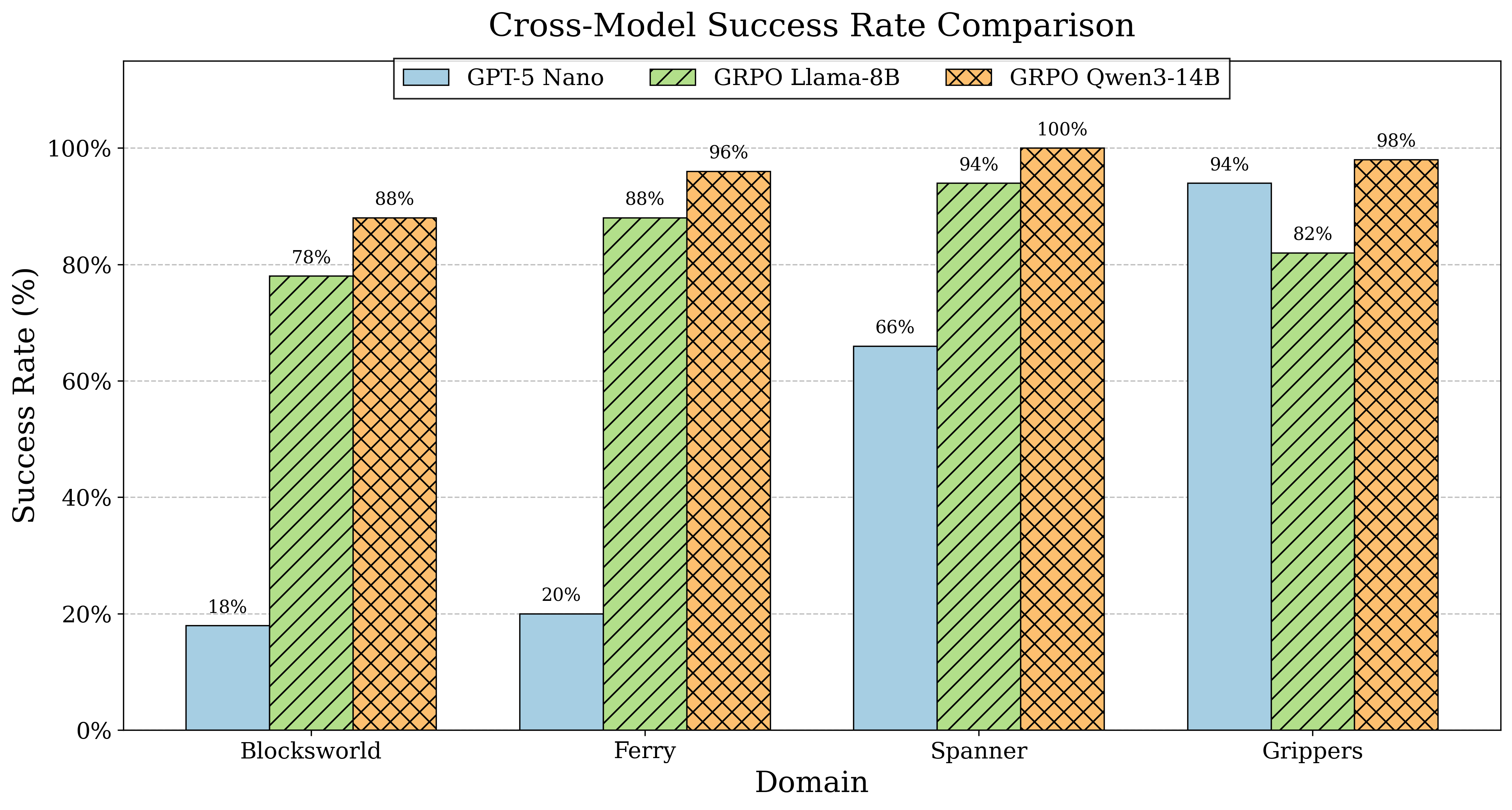}
  \caption{Cross-model success rate comparison across four domains.}
  \label{fig:cross-model-success}
\end{figure}

To further contextualize these results, we compare our GRPO-trained models against GPT-5 Nano\footnote{\texttt{gpt-5-nano-2025-08-07}, evaluated via the OpenAI API with the same prompt template used for our models in a zero-shot setting.}~\cite{openai2025gpt5}.
Figure~\ref{fig:cross-model-success} presents the success rates across four domains.
GRPO Qwen3-14B achieves the highest success across all domains (88--100\%), followed by GRPO Llama-8B (78--94\%).
In contrast, GPT-5 Nano achieves only 18\% and 20\% on Blocksworld and Ferry, respectively, despite strong performance on Grippers (94\%). These results demonstrate that targeted fine-tuning with verifiable safety rewards enables smaller open-source models to surpass much larger proprietary models on safety-constrained planning tasks.

\subsection{Input Format Comparison: Symbolic vs Natural Language vs JSON}

To evaluate the robustness of our framework to different input representations, we compare model performance across three input formats: symbolic PDDL3, natural language (NL), and JSON.
For NL, we convert PDDL3 specifications into human-readable prompts; for JSON, we use a structured key-value representation that preserves the semantic content while providing a machine-friendly format.

Table~\ref{tab:input_format_comparison} reports the results across four domains.
Although the model is trained exclusively on PDDL3 input, it achieves strong performance on both NL and JSON formats with near-zero format errors.
JSON achieves 92.5\% average success rate across the four standard domains, matching PDDL3 in Ferry (96.0\%) and Grippers (98.0\%).
NL also performs well with 84.0\% average success, demonstrating that the model can interpret human-readable problem descriptions.
Notably, format errors remain negligible across all three formats (0.0\% for PDDL3 and JSON, 0.5\% for NL), indicating that the learned planning knowledge transfers effectively to unseen input representations.
These results confirm that our GRPO-trained model generalizes beyond its input representation, enabling flexible deployment with different input interfaces.

\subsection{Integration with LLM Agentic Workflows}

A key advantage of our approach is that the trained model can be seamlessly integrated with existing LLM agentic workflows to further improve reliability.
We evaluate this by combining our GRPO-trained model with SafePilot~\cite{xu2024assuring}, a verification-guided iterative refinement framework that detects plan errors and prompts the LLM to regenerate until a valid plan is produced.

\begin{table}[t]
\centering
\caption{Comparison of Planning Success Rates and Retry Counts with SafePilot.}
\label{tab:safepilot_comparison}
\resizebox{\columnwidth}{!}{%
\begin{tabular}{lccccc}
\toprule
\textbf{Model} & \textbf{Blocksworld} & \textbf{Ferry} & \textbf{Grippers} & \textbf{Spanner} & \textbf{Avg.} \\
\midrule
\multicolumn{6}{l}{\textit{Success Rate (\%) $\uparrow$}} \\
Pretrained & 50.0 & 0.0 & 50.0 & 0.0 & \textbf{25.0} \\
GRPO & 94.0 & 98.0 & 98.0 & 100.0 & \textbf{97.5} \\
\midrule
\multicolumn{6}{l}{\textit{Avg. Retries $\downarrow$}} \\
Pretrained & 3.50 & 5.00 & 3.00 & 5.00 & \textbf{4.13} \\
GRPO & 1.36 & 1.12 & 1.14 & 1.00 & \textbf{1.16} \\
\bottomrule
\end{tabular}%
}
\end{table}

Table~\ref{tab:safepilot_comparison} presents the results of combining both the Pretrained and GRPO-trained Qwen3-14B models with SafePilot.
We highlight two key findings:

\paragraph{GRPO Training Is Essential for SafePilot Effectiveness}
Even with SafePilot's iterative refinement, the Pretrained model achieves only 25.0\% average success with 4.13 retries per problem, failing entirely on Ferry and Spanner.
In contrast, the GRPO model reaches 97.5\% average success with just 1.16 retries, attaining 100\% on Spanner with single-pass generation.
This demonstrates that SafePilot alone cannot compensate for a weak base model---GRPO training is essential for internalizing domain knowledge and safety constraints.

\paragraph{Seamless Integration with LLM Workflows.}
Compared to the standalone GRPO result of 95.5\% (Table~\ref{tab:error_percentages}), SafePilot provides a modest but meaningful boost to 97.5\%, catching residual errors with minimal overhead (1.16 retries on average).
This confirms that our fine-tuned model integrates seamlessly with existing LLM agentic workflows, and that the combination of internalized safety knowledge and external verification yields a highly reliable planning system.
\begin{figure}[!t]
  \centering
  \begin{subfigure}[b]{\linewidth}
    \centering
    \includegraphics[width=0.9\linewidth]{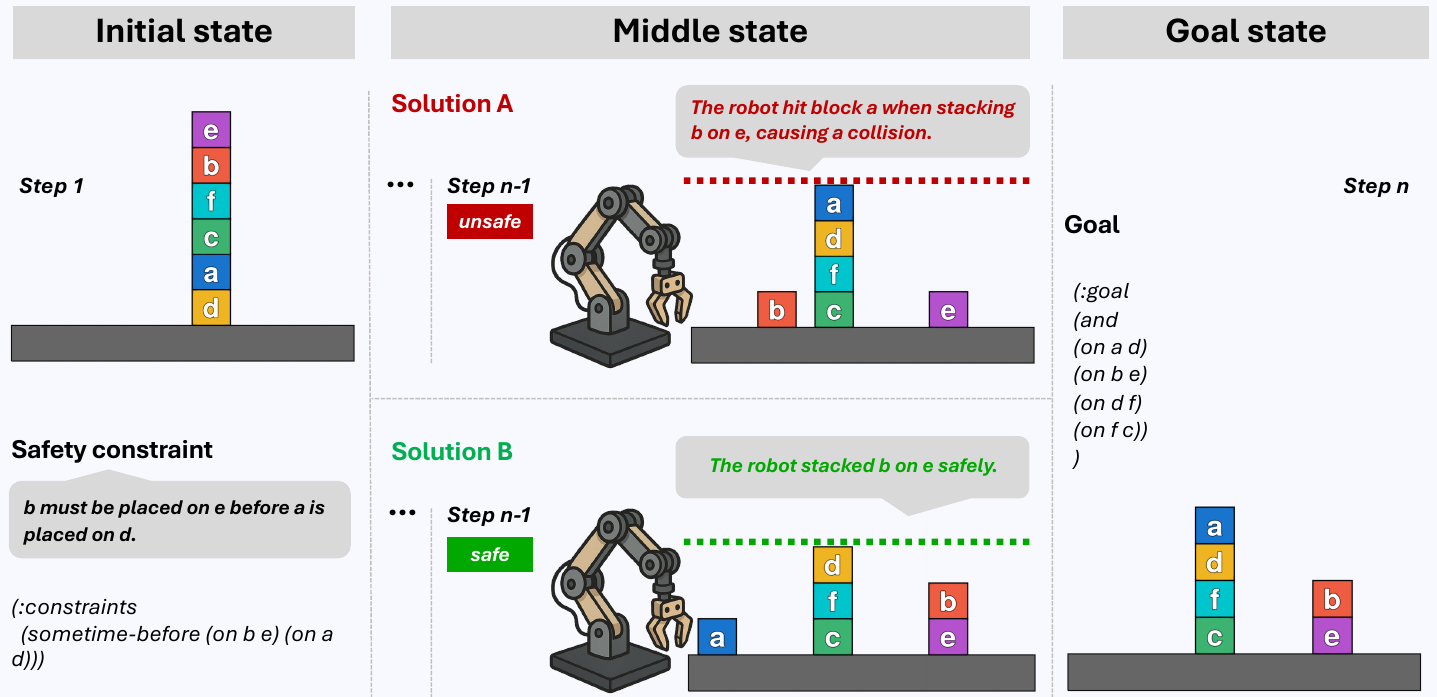}
    \caption{Case study in the Blocksworld domain.}
    \label{fig:case-study}
  \end{subfigure}
  \vspace{0.5em}
  \begin{subfigure}[b]{\linewidth}
    \centering
    \includegraphics[width=0.9\linewidth]{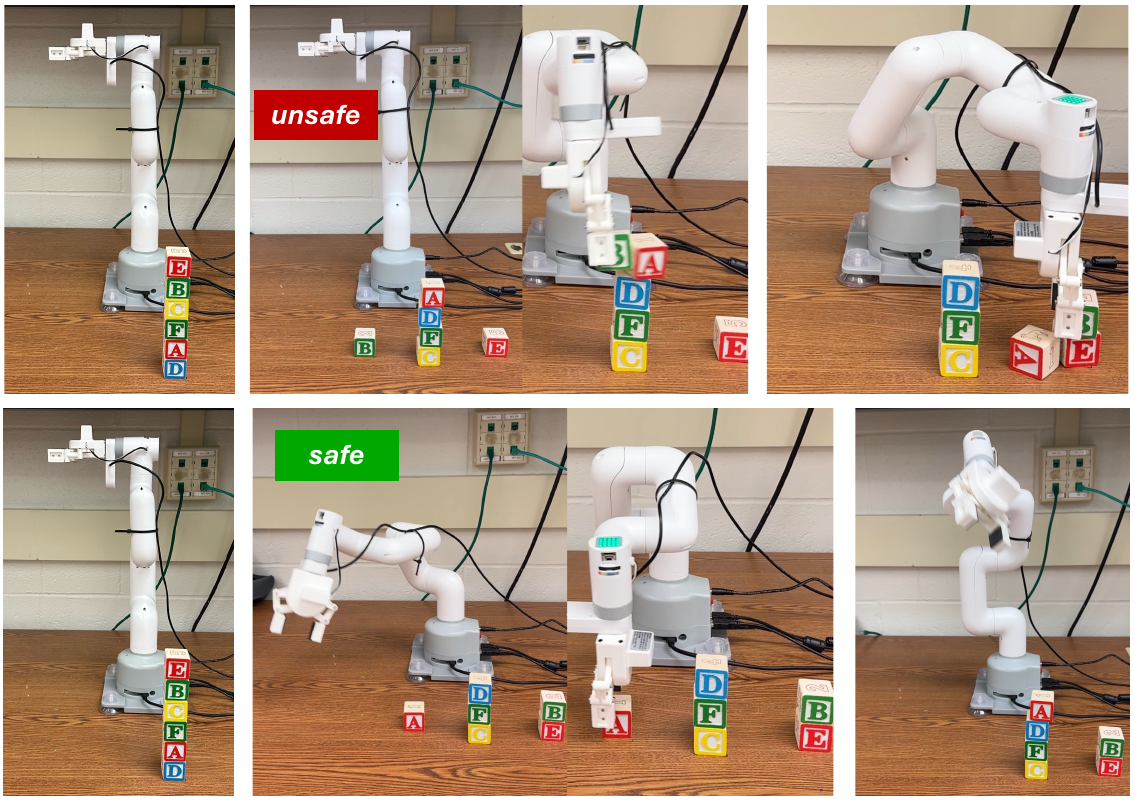}
    \caption{Snapshot of the physical robot executing the Blocksworld task.}
    \label{fig:experiment-snapshot}
  \end{subfigure}
  \caption{Real-world validation in Blocksworld: (a) simulation case study comparing a classical solver (unsafe) with our safety-aware planner (safe); (b) physical deployment on a robot arm.}
  \label{fig:real-world-validation}
\end{figure}

\subsection{Real-World Validation in Blocksworld}

To demonstrate the practical impact of safety-aware training, we evaluate our framework in both simulation and a physical robotic system setup using the classical Blocksworld domain.

\paragraph{Simulation.}
We first select a test-set problem instance and compare two planners:
(i) a classical PDDL2 solver, and
(ii) our safety-aware SFT+GRPO model.
The task requires satisfying a safety constraint that enforces a specific ordering between two stacking operations to avoid unsafe intermediate configurations.

As shown in Figure~\ref{fig:case-study}, the classical solver produces a plan that violates the constraint, leading to an intermediate configuration that would cause a collision between blocks.
In contrast, the safety-aware LLM planner restructures the action sequence so that the constraint is satisfied while still achieving the goal, demonstrating that the learned safety knowledge directly influences plan generation.

\paragraph{Physical Deployment.}
We further deploy the safety-aware planner on a desktop robot arm (Elephant myCobot~280) controlled by a Raspberry~Pi.
The LLM-generated plan is transmitted via SSH and executed on the real robot.
We compare the safety-aware plan with a baseline unsafe sequence that mirrors the violation observed in simulation.

Figure~\ref{fig:real-world-validation}(b) shows snapshots from the execution.
The unsafe baseline results in a physical collision during stacking, whereas the safety-aware plan completes the task without violating the safety requirement.
This experiment confirms that the proposed framework not only improves symbolic safety in simulation but also yields robust, collision-free behaviors when deployed on real robotic hardware.

\section{Conclusion}\label{VI}
We presented a two-stage post-training framework that combines SFT with GRPO to enable LLMs to perform safety-aware task planning for robotic systems.
By injecting verifiable safety knowledge through online RL with fine-grained reward machines, our approach achieves high planning success rates with minimal safety violations, and generalizes to unseen problems across a set of training domains. Since our experiments use quantized lightweight models and a moderately sized dataset, the reported performance is likely conservative.
Future work includes scaling to more complex robotic settings, integrating richer formal verification tools, and exploring automatic construction of safety constraints from interaction data.

\bibliographystyle{IEEEtran}
\bibliography{sample-base}
\newpage
\appendix

\subsection{Training Configurations}\label{app:training-details}

All models are fine-tuned on a single NVIDIA H100 GPU with 80~GB memory.
Our implementation is based on Python~3.10 and PyTorch~2.8.0~\cite{paszke2019pytorch}, and uses the Hugging Face Transformers library~\cite{wolf2020transformers}, TRL~\cite{vonwerra2022trl}, and Unsloth~\cite{unsloth}.
We adopt Low-Rank Adaptation (LoRA)~\cite{hu2022lora} and Quantized LoRA (QLoRA)~\cite{dettmers2023qlora} as parameter-efficient fine-tuning strategies to reduce memory usage and training time.
Table~\ref{tab:hyperparameters} summarizes the key hyperparameters for both training stages.
Tables~\ref{tab:reward-config} and~\ref{tab:curriculum-config} detail the reward function and curriculum learning settings used in GRPO training.

\begin{table}[h]
  \centering
  \caption{Training hyperparameters for SFT and GRPO.}
  \label{tab:hyperparameters}
  \resizebox{\columnwidth}{!}{%
  \begin{tabular}{lcc}
    \hline
    \textbf{Hyperparameter} & \textbf{SFT} & \textbf{GRPO} \\
    \hline
    Learning rate         & $2 \times 10^{-4}$  & $1 \times 10^{-5}$ \\
    LR scheduler          & Cosine               & Cosine \\
    Batch size            & 4                    & 8 \\
    Gradient accumulation & 2                    & 4 \\
    Effective batch size  & 8                    & 32 \\
    Training duration     & 3 epochs             & 1{,}000 steps \\
    Warmup ratio          & 0.1                  & --- \\
    Weight decay          & 0.05                 & --- \\
    KL penalty ($\beta$)  & ---                  & 0.01 \\
    Generations per prompt ($K$) & ---            & 8 \\
    Sampling temperature  & ---                  & 0.6 \\
    LoRA rank / alpha     & 32 / 64              & 32 / 64 \\
    LoRA dropout          & 0.05                 & 0.05 \\
    Max sequence length   & 4{,}096              & 2{,}048 \\
    Precision             & bfloat16             & bfloat16 \\
    \hline
  \end{tabular}%
  }
\end{table}

\subsection{Reward and Curriculum Settings}\label{app:reward-curriculum}

\begin{table}[h]
  \centering
  \caption{Reward intervals for each validation category.}
  \label{tab:reward-config}
  \begin{tabular}{lc}
    \hline
    \textbf{Category} & \textbf{Reward range} \\
    \hline
    Success ($c_5$)                 & $+1.0$ (fixed) \\
    Goal not satisfied ($c_4$)      & $[-0.4,\;-0.1]$ \\
    Precondition violation ($c_3$)  & $[-0.6,\;-0.3]$ \\
    Safety violation ($c_2$)        & $[-0.9,\;-0.6]$ \\
    Format error ($c_1$)            & $-1.0$ (fixed) \\
    \hline
  \end{tabular}
\end{table}

\begin{table}[h]
  \centering
  \caption{Curriculum schedule: sampling probabilities by training phase.}
  \label{tab:curriculum-config}
  \begin{tabular}{lcccc}
    \hline
    \textbf{Phase} & \textbf{Progress} & \textbf{Easy} & \textbf{Medium} & \textbf{Hard} \\
    \hline
    Early & 0--30\%   & 70\% & 25\% & 5\% \\
    Mid   & 30--70\%  & 40\% & 40\% & 20\% \\
    Late  & 70--100\% & 20\% & 40\% & 40\% \\
    \hline
  \end{tabular}
\end{table}

\subsection{Dataset Statistics}\label{app:dataset-stats}

Table~\ref{tab:dataset-stats} summarizes the dataset statistics for each domain.
For each domain, we generate 500 training problems (used for both SFT and GRPO) and 50 test problems, all solved by OPTIC and validated by VAL.
Problem complexity varies within each domain based on the number of objects and structural parameters.

\begin{table}[h]
  \centering
  \caption{Dataset statistics and problem size ranges per domain.}
  \label{tab:dataset-stats}
  \resizebox{\columnwidth}{!}{%
  \begin{tabular}{lccl}
    \hline
    \textbf{Domain} & \textbf{Train} & \textbf{Test} & \textbf{Problem size range} \\
    \hline
    Blocksworld & 500 & 50 & 3--6 blocks \\
    Ferry       & 500 & 50 & 3--4 locations, 2--3 cars \\
    Grippers    & 500 & 50 & 1 robot, 3--4 rooms, 3 objects \\
    Spanner     & 500 & 50 & 2--3 spanners, 2 nuts, 3--4 locations \\
    \hline
    \textbf{Total} & \textbf{2{,}000} & \textbf{200} & --- \\
    \hline
  \end{tabular}%
  }
\end{table}

\subsection{PDDL3 Constraint Examples}\label{app:constraint-examples}

Each domain uses distinct PDDL3 constraint types that encode domain-specific safety requirements.
Below we provide a representative constraint instance from each domain's test set.

\noindent\textbf{Blocksworld} uses \texttt{sometime-before} constraints to enforce safe stacking order:
\begin{lstlisting}[style=lispstyle]
(:constraints
  (sometime-before (on-table b2)
                   (on-table b1)))
\end{lstlisting}
This requires that block~\texttt{b1} must be placed on the table at some point \emph{before} block~\texttt{b2} is placed on the table, preventing unsafe intermediate configurations.

\noindent\textbf{Ferry} uses \texttt{sometime-before} constraints to enforce safe transport ordering:
\begin{lstlisting}[style=lispstyle]
(:constraints
  (and
    (sometime-before (at c0 l0)
                     (at-ferry l2))
    (sometime-before (at c1 l0)
                     (at-ferry l1))))
\end{lstlisting}
This requires the ferry to visit specific locations before delivering each car to its destination.

\noindent\textbf{Spanner} uses compound \texttt{always-imply} and \texttt{at-most-once} constraints with quantifiers:
\begin{lstlisting}[style=lispstyle]
(:constraints
  (and
    (always (imply (not (tightened nut1))
                   (not (tightened nut2))))
    (forall (?m - man)
      (at-most-once (at ?m shed)))))
\end{lstlisting}
The first constraint enforces a tightening order (nut2 cannot be tightened before nut1), and the second restricts each worker to enter the tool shed at most once.

\noindent\textbf{Grippers} uses \texttt{always} constraints with universal quantification:
\begin{lstlisting}[style=lispstyle]
(:constraints
  (always (forall (?b - object)
    (not (carry robot1 ?b rgripper1)))))
\end{lstlisting}
This prohibits a specific gripper from ever carrying any object throughout the entire plan, modeling a safety restriction on a malfunctioning or reserved gripper.

\subsection{Validation Pipeline}\label{app:validation-pipeline}

We use the VAL tool~\cite{howey2004val} to validate generated plans.
Each plan is written to a temporary file and validated by running \texttt{Validate -v <domain> <problem> <plan>} with a 30-second timeout.
The validation output is then classified into one of five categories based on pattern matching on the validator's stdout:

\begin{enumerate}
  \item \textbf{Success}: Output contains \texttt{"Plan valid"} --- the plan satisfies all preconditions, safety constraints, and goals.
  \item \textbf{Plan format error}: Output contains indicators such as \texttt{"bad operator in plan"}, \texttt{"no matching action defined"}, or \texttt{"type problem with proposition"} --- the plan is syntactically malformed or references undefined actions/objects.
  \item \textbf{Goal not satisfied}: Output contains \texttt{"goal not satisfied"} --- the plan executes without errors but does not achieve the goal state.
  \item \textbf{Precondition violation}: Output contains \texttt{"plan failed to execute"} together with \texttt{"unsatisfied precondition"} --- an action's preconditions are not met at the point of execution.
  \item \textbf{Safety constraint violation}: Output contains \texttt{"plan failed to execute"} without any precondition failure --- a PDDL3 constraint is violated during execution.
\end{enumerate}

This classification is used both for evaluation metrics (Section~\ref{V}) and for computing GRPO rewards during training (Section~\ref{subsubsec:reward}).

\subsection{Evaluation Configuration}\label{app:eval-config}

Table~\ref{tab:eval-config} summarizes the evaluation parameters for both local models and API-based models.

\begin{table}[h]
  \centering
  \caption{Evaluation parameters for local and API-based models.}
  \label{tab:eval-config}
  \begin{tabular}{lcc}
    \hline
    \textbf{Parameter} & \textbf{Local} & \textbf{API} \\
    \hline
    Temperature         & 0.6      & default \\
    Top-$p$             & 0.9      & default \\
    Max new tokens      & 1{,}024  & default \\
    Max sequence length  & 5{,}000  & --- \\
    Quantization        & 4-bit    & --- \\
    Timeout per problem & ---      & 300s \\
    Max retries         & ---      & 5 \\
    Concurrent threads  & ---      & 5 \\
    \hline
  \end{tabular}
\end{table}

For local models, we use greedy-like sampling with temperature~0.6 and top-$p$~0.9.
API-based models (e.g., GPT-5 Nano) use their default generation settings.
All models receive the same zero-shot prompt template described in Section~\ref{subsec:dataset}.

\subsection{Input Format Conversion}\label{app:input-format}

To evaluate robustness to different input representations (Section~\ref{V}), we convert PDDL3 problem files into natural language (NL) and JSON formats.

\noindent\textbf{Natural language conversion.}
Each PDDL3 problem is converted to a structured NL description using domain-specific templates.
The NL output includes sections for objects, initial state, goal, and constraints.
For example, the Blocksworld predicate \texttt{(on b3 b2)} is rendered as ``Block b3 is on block b2,'' and the constraint \texttt{(sometime-before (on-table b2) (on-table b1))} becomes ``Before `block b2 is on the table' becomes true, `block b1 is on the table' must be true at some point.''
All PDDL3 constraint types (\texttt{sometime-before}, \texttt{always}, \texttt{at-most-once}, \texttt{forall}, etc.) have corresponding NL templates.

\noindent\textbf{JSON conversion.}
The JSON format preserves the semantic structure of PDDL3 using key-value pairs.
Objects, initial state predicates, goal conditions, and constraints are represented as structured fields.
For example, an initial state predicate \texttt{(on b3 b2)} becomes \texttt{\{"pred": "on", "args": ["b3", "b2"]\}}, and constraints are stored with their type as the key (e.g., \texttt{"sometime\_before": [\ldots]}).

Both formats are generated automatically from the original PDDL3 files and validated by ensuring that the corresponding PDDL3 solution file remains applicable.

\end{document}